\documentclass[letterpaper]{article} 
\usepackage{aaai23}  
\usepackage{times}  
\usepackage{helvet}  
\usepackage{courier}  
\usepackage[hyphens]{url}  
\usepackage{graphicx} 
\usepackage{natbib}  
\usepackage{caption} 
\usepackage{times}
\usepackage{latexsym}
\usepackage{verbatim}
\usepackage{amsmath}
\usepackage{multirow}
\usepackage{booktabs}
\usepackage{tabu}

\usepackage{algorithm}
\usepackage{algorithmic}

\usepackage{newfloat}
\usepackage{listings}
\DeclareCaptionStyle{ruled}{labelfont=normalfont,labelsep=colon,strut=off} 
\floatstyle{ruled}
\newfloat{listing}{tb}{lst}{}
\floatname{listing}{Listing}
\title{DHGE: Dual-View Hyper-Relational Knowledge Graph Embedding for Link Prediction and Entity Typing}
\author{
    Haoran Luo,
    Haihong E\thanks{Corresponding author.},
    Ling Tan,
    Gengxian Zhou,
    Tianyu Yao,
    Kaiyang Wan
}
\affiliations{
    School of Computer Science, Beijing University of Posts and Telecommunications, Beijing, China\\
    \{luohaoran, ehaihong\}@bupt.edu.cn
}

\usepackage{bibentry}

\begin{document}

\maketitle

\begin{abstract}
In the field of representation learning on knowledge graphs (KGs), a hyper-relational fact consists of a main triple and several auxiliary attribute-value descriptions, which is considered more comprehensive and specific than a triple-based fact. However, currently available hyper-relational KG embedding methods in a single view are limited in application because they weaken the hierarchical structure that represents the affiliation between entities. To overcome this limitation, we propose a dual-view hyper-relational KG structure (DH-KG) that contains a hyper-relational instance view for entities and a hyper-relational ontology view for concepts that are abstracted hierarchically from the entities. This paper defines link prediction and entity typing tasks on DH-KG for the first time and constructs two DH-KG datasets, JW44K-6K, extracted from Wikidata, and HTDM based on medical data. Furthermore, we propose DHGE, a DH-KG embedding model based on GRAN encoders, HGNNs, and joint learning. DHGE outperforms baseline models on DH-KG, according to experimental results. Finally, we provide an example of how this technology can be used to treat hypertension. Our model and new datasets are publicly available.
\end{abstract}

\section{Introduction}
\label{s1}

Modern large-scale knowledge graphs (KGs), such as Freebase~\cite{Freebase} and Wikidata~\cite{Wikidata}, consist of facts with entities and relations. A triple-based fact has two entities and one relation, which is usually expressed by triples (\textit{subject}, \textit{relation}, \textit{object}).

\begin{figure}[h!t]
    \centering
    \includegraphics[width=7.5cm]{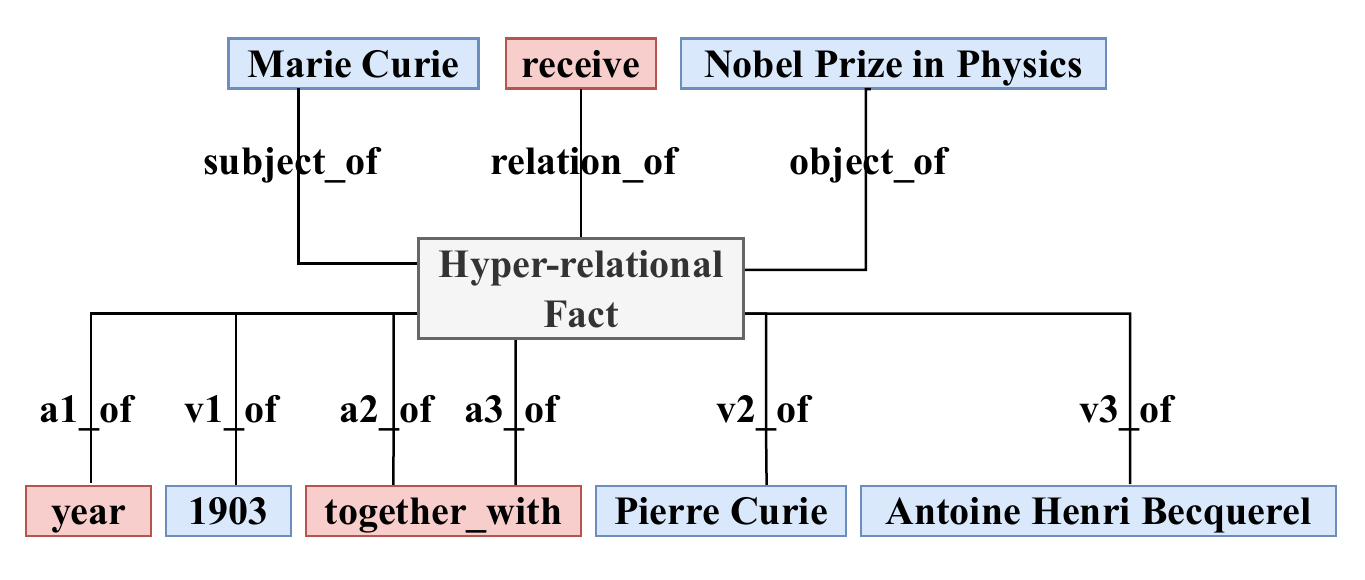}
    \caption{Hyper-relational Structure.}
    \label{f1}
\end{figure}

Recent studies~\citep{m-TransH, NaLP} suggest that some facts contain more than two entities, which are represented as hyper-relational facts~\citep{HINGE}. In contrast to triples, hyper-relational facts consist of one main triple (\textit{s},\textit{r},\textit{o}) and several auxiliary attribute-value descriptions \{(\textit{a$_i$ }:\textit{v$_i$})\}. For example, the fact that \textit{Marie Curie received Nobel Prize in Physics in 1903 together with Pierre Curie and Antoine Henri Becquerel}, can be expressed as a hyper-relational fact: (\textit{Marie Curie}, \textit{receive}, \textit{Nobel Prize in Physics}, \textit{year}, \textit{1903}, \textit{together\_with}, \textit{Pierre Curie}, \textit{together\_with}, \textit{Antoine Henri Becquerel}) as shown in Figure~\ref{f1}.

\begin{figure}[h!t]
    \centering
    \includegraphics[width=7.5cm]{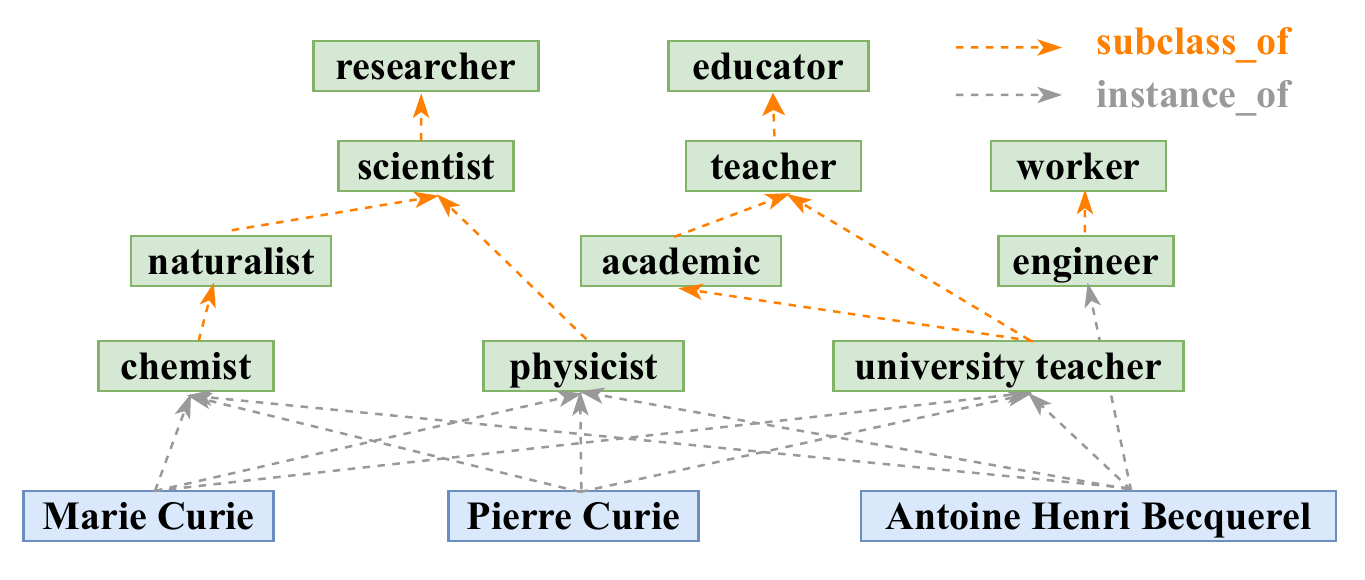}
    \caption{Hierarchical Structure.}
    \label{f2}
\end{figure}

\begin{figure*}[h!t]
    \centering
    \includegraphics[width=16cm]{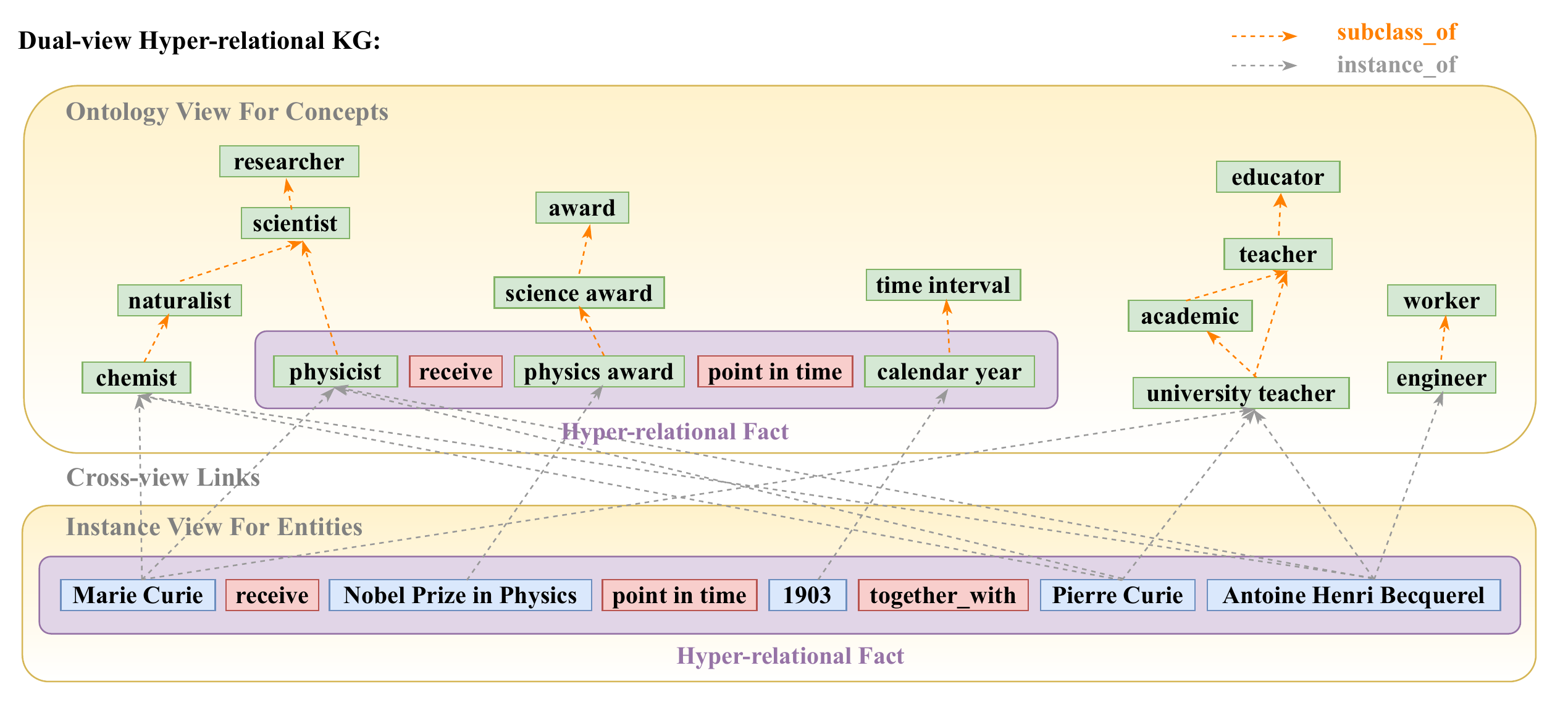}
    \caption{An example on Dual-view Hyper-Relational KG Structure, where a hyper-relational fact in the instance view consists of a main triple (\textit{Marie Curie}, \textit{receive}, \textit{Nobel Prize in Physics}) and auxiliary attribute-value descriptions (\textit{year}: \textit{1903}, \textit{together\_with}: \textit{Pierre Curie}, \textit{together\_with}: \textit{Antoine Henri Becquerel}). Additionally, entities in instance view and concepts in ontology view are linked across views through the ``\texttt{instance\_of}" relation, and many concepts in the ontology view are connected by the ``\texttt{subclass\_of}" relation.}
    \label{f3}
\end{figure*}

However, the hyper-relational structure weakens the hierarchical structure of KGs due to the consideration of auxiliary information. As pointed out by \citet{KACC}, the KG hierarchical structure consists of triple-based facts with hierarchical relations such as ``\texttt{instance\_of}" and ``\texttt{subclass\_of}". As shown in Figure~\ref{f2}, (\textit{Marie Curie}, \texttt{instance\_of}, \textit{chemist}) represents the affiliation between the entity \textit{Marie Curie} and the concept \textit{chemist}, and (\textit{chemist}, \texttt{subclass\_of}, \textit{naturalist}) represents the affiliation between the concept \textit{chemist} and the more abstract concept \textit{naturalist}. 

In real-world KGs, hierarchical and hyper relations are essential in describing knowledge facts. Despite this, no research has been conducted on the joint modeling of the two relations, severely limiting the application of the hyper-relational KG embedding methods. In our work, we are primarily concerned with overcoming this limitation.

For the first time, this paper defines the structure of hyper-relational dual-view KG (DH-KG), as Figure~\ref{f3} shows, consisting of a hyper-relational instance view for entities and a hyper-relational ontology view for concepts abstracted hierarchically from entities. The two views are connected by cross-view links, which link entities and corresponding concepts. Based on this structure, we construct two DH-KG datasets, JW44K-6K extracted from Wikidata~\citep{Wikidata}, and HTDM based on medical data about hypertension and describe their construction method.

We also propose DHGE, the first model that can jointly model both hyper-relational facts and dual-view hierarchical information. DHGE utilizes the GRAN encoder~\citep{GRAN}, hypergraph neural network~\citep{HGNN}, and joint learning to perform the hyper-relational KG link prediction task within two views and the cross-view linked entity typing task. We conduct extensive experiments comparing our model with other models and analyze the results. Our model performs well on JW44K-6K. Additionally, we provide an example of HTDM application in hypertension.

Our main contributions are summarized as follows:
\begin{itemize}
\item We define the structure of DH-KG and its link prediction and entity typing tasks for the first time.
\item We construct two DH-KG datasets, JW44K-6K and HTDM.
\item We propose DHGE, the first DH-KG embedding model that can simultaneously model hyper-relational facts and hierarchical information. 
\item We conduct extensive experiments and comparative analyses to verify our model's good performance on DH-KGs. 
\item We provide an example of the application in the field of hypertension on HTDM. 
\end{itemize}

\begin{table*}[t]
\small
\centering
\label{tab4}
\setlength{\tabcolsep}{0.03mm}{
\begin{tabular}{llccccccccccc}
\toprule \multirow{2.5}{*}{\textbf{Category}} &\multirow{2.5}{*}{\textbf{Dataset}} & \multicolumn{5}{c}{ \textbf{The Instance View} } & \multicolumn{5}{c}{ \textbf{The Ontology View} }& \multirow{2.5}{*}{\textbf{\# Cross-links}} \\
\cmidrule(lr){3-7}\cmidrule(lr){8-12}&& \textbf{\# Entities} & \textbf{\# Rels} & \textbf{\# Facts} &  \textbf{Arity} & \textbf{\#[A\textgreater2]Facts(\%)} & \textbf{\# Concepts} & \textbf{\# Rels} & \textbf{\# Facts} &  \textbf{Arity} &\textbf{\#[A\textgreater 2]Facts(\%)}  \\
\midrule  \multirow{4}{*}{\textbf{ST-KGs}}
& FB15K & 14,951 & 1,345 & 582,213 &2&-&-&-&-&-&-&- \\
& FB15K-237 & 14,541 & 237 & 310,116 &2&-&-&-&-&-&-&- \\
& WN18 & 40,943 & 18 & 151,442 &2&-&-&-&-&-&-&- \\
& WN18RR & 40,943 & 11 & 93,003 &2&-&-&-&-&-&-&- \\
\midrule  \multirow{3}{*}{\textbf{SH-KGs}}
& JF17K & 28,645 & 322 & 100,947 &2-6& 46,320(45.9\%) &-&-&-&-&-&- \\
& Wikipeople & 34,839 &  375 &  369,866 &2-9& 9,482(2.6\%) &-&-&-&-&-&- \\
& WD50K & 47,156 &  532 & 236,507 &2-7& 32,167(13.6\%) &-&-&-&-&-&- \\
\midrule  \multirow{3}{*}{\textbf{DT-KGs}}
& YAGO26K-906 & 26,078 & 34 & 390,738 &2&-& 906 & 30 & 8,962 &2&-& 9,962 \\
& DB111K-174 & 111,762 & 305 & 863,643 &2&-& 174 & 20 & 763 &2&-& 99,748 \\
& KACC-M & 99,615 & 209 & 662,650 &2&-& 6,685 & 30 & 15,616 &2&-& 123,342 \\
\midrule\midrule  \multirow{2}{*}{\textbf{DH-KGs}}
& JW44K-6K & 44,342 & 630 & 514,822 &2-10&  50,869(9.9\%) & 6,589 & 141 & 17,667 &2-7& 155(0.9\%) & 61,874 \\
& HTDM & 1,297 & 45 & 208 &7-45& 208(100\%) & 266 & 19 & 470&2-10& 296(63.0\%) & 469 \\
\bottomrule
\end{tabular}}
\caption{\label{T1}
Data statistics of datasets of ST-KGs (Single-view Triple-based KGs), SH-KGs (Single-view Hyper-relational KGs), DT-KGs (Dual-view Triple-based KGs), and DH-KGs (Dual-view Hyper-relational KGs), where the columns indicate the number of entities, relations, facts, arity of facts, and facts with more than two arities, respectively.
}
\end{table*}

\section{Related Work}
\label{s2}
\subsection{Knowledge Graph Datasets}
Knowledge graph (KG) datasets are generally subgraphs extracted from large KGs, which can be summarized in the following four forms.

{\bf Single-view Triple-based KG Datasets. }
The original KG datasets are in the form of single-view triples such as FB15K, FB15K-237~\citep{TransE} from Freebase, and WN18,  WN18RR~\citep{TransE} from Wordnet.

{\bf Single-view Hyper-relational KG Datasets. }
To represent the hyper-relational facts, JF17K~\citep{m-TransH}, WikiPeople~\citep{NaLP}, WD50K~\citep{StarE}, and other datasets extend the triplet paradigm to a hyper-relational paradigm but still in a single view.

{\bf Dual-view Triple-based KG Datasets. }
To better consider the hierarchical information between entities and concepts, YAGO26K-906, DB111K-174~\citep{JOIE}, and KACC~\citep{KACC} adopt the dual-view structure, but they are still limited to the triple form.

{\bf Dual-view Hyper-relational KG Datasets. }
Our proposed DH-KG datasets, JW44K-6K and HTDM, contain high-quality instance and ontology views with hyper-relational facts and are rich in cross-view links. To our best knowledge, JW44K-6K is currently the only DH-KG dataset closest to real-world facts, while HTDM is the only DH-KG dataset for medical applications.

\subsection{Knowledge Graph Embedding Methods}
The knowledge graph embedding (KGE) methods determine the performance of the knowledge representation, which can be summarized in the following four categories.

{\bf Single-view Triple-based KGE Methods. }
Traditional KGE methods mainly model triples on a single view, such as translation models~\citep{TransE, RotatE}, tensor factorization base models~\citep{TuckER}, and neural models~\citep{ConvE}.

{\bf Single-view Hyper-relational KGE Methods. }
Some methods, such as m-TransH~\citep{m-TransH} and NaLP~\citep{NaLP}, extend the triple-based paradigm with binary relations to multiple relations. HINGE~\citep{HINGE}, StarE~\citep{StarE}, and GRAN~\citep{GRAN} utilize a hyper-relational paradigm to model multiple relations in a single view.

{\bf Dual-view Triple-based KGE Methods. }
Models such as JOIE~\citep{JOIE} and HyperKA~\citep{HyperKA} can jointly model the information in dual-view to simultaneously model hierarchical and logical information, but they still cannot model hyper-relational facts.

{\bf Dual-view Hyper-relational KGE Methods. }
Our proposed model, DHGE, extends hyper-relational KG embedding to a dual-view structure. To our best knowledge, DHGE is currently the only KGE model that can jointly model dual-view hierarchical relations and hyper-relational facts.

\section{Problem Statement}
\label{s3}
In this section, we present the formalization of dual-view hyper-relational KG (DH-KG) and the definition of link prediction and entity typing tasks.

\textbf{Formalization of DH-KG.} DH-KG contains two sub-views and some cross-view links, and we denote it as $\{\mathcal{G}_I, \mathcal{G}_O, \mathcal{H}_S\}$. The instance view $\mathcal{G}_I=\{\mathcal{E}_I, \mathcal{R}_I, \mathcal{H}_I\}$ consists of an instance entity set $\mathcal{E}_I$, an instance relation set $\mathcal{R}_I$, and an instance hyper-relational fact set $\mathcal{H}_I=\{((s^I,r^I,o^I),\{(a^I_i:v^I_i)\}^m_{i=1})|s^I,o^I,v^I_1, ...,v^I_m\in\mathcal{E}_I, r^I,a^I_1,...,a^I_m\in\mathcal{R}_I\}$, where $(s,r,o)$ represents the main triple and $\{(a^I_i:v^I_i)\}^m_{i=1}$ represents m auxiliary attribute-value pairs. Similarly, the ontology view $\mathcal{G}_O=\{\mathcal{C}_O,\mathcal{R}_O,\mathcal{H}_O\}$ consists of an ontology concept set $\mathcal{C}_O$, an ontology relation set $\mathcal{R}_O$, and an ontology hyper-relational fact set $\mathcal{H}_O=\{((s^O,r^O,o^O),\{(a^O_i:v^O_i)\}^m_{i=1})|s^O,o^O,v^O_1,...,v^O_m\in\mathcal{C}_O, r^O,a^O_1,...,a^O_m\in\mathcal{R}_O\}$. The set of cross-view links $\mathcal{H}_S=\{(h_S,\texttt{instance\_of},t_S)\}$ is a set of hyper-relational facts without auxiliary attribute-value pairs, where $h_S \in\mathcal{E}_I$ and $t_S\in\mathcal{C}_O$.

\textbf{Link prediction (LP) on DH-KG.} The hyper-relational LP task aims to predict missing elements from hyper-relation facts$((s^I,r^I,o^I),\{(a_i:v_i)\}^m_{i=1})$. Missing elements can be entities $\in\{s,o,v_1,...,v_m\}$ or relations $\in\{r,a_1,...,a_m\}$. On DH-KG, $\mathcal{G}_I$ LP and $\mathcal{G}_O$ LP are hyper-relational LP in the instance and ontology view, respectively.

\textbf{Entity typing (ET) on DH-KG.} The ET task attempts to predict the associated concepts of some given entity. On DH-KG, the ET task is to predict the tail of $(h,\texttt{instance\_of},t)$ in cross-view links, where $h\in\mathcal{E}_I$, $t\in\mathcal{C}_O$.

\section{Dataset Construction}
\label{s4}
In this section, we introduce the construction method, data statistics, and analysis of two DH-KG datasets\footnote{\url{https://github.com/LHRLAB/DHGE/tree/main/dataset}}, JW44K-6K and HTDM.

\subsection{JW44K-6K Dataset}
JW44K-6K is a DH-KG dataset extracted from Wikidata~\citep{Wikidata}. It consists of 44K instance-view entities and 6K ontology-view concepts, with rich cross-view links and intra-view hyper-relational facts. The construction of the JW44K-6K dataset can be divided into four steps.

\textbf{Step1. Instance-view entity filtering.} 
A hyper-relational KG dataset WD50K, proposed by \citealp{StarE}, is extracted from Wikidata, containing 50K entities. We take the entities of WD50K as the instance-view entity set $\mathcal{E}_I$, then take $\mathcal{E}_I$ as the head entity set to find the corresponding tail entity set $T$ through the relation ``\texttt{instance\_of}" in Wikidata, and update $\mathcal{E}_I$ to $\mathcal{E}_I-\mathcal{E}_I \cap T$, thereby filtering out conceptual entities. The filtered $\mathcal{E}_I$ has 44K entities.

\textbf{Step2. Ontology-view concept selection.} 
We continue to find the tail entity set corresponding to $\mathcal{E}_I$ through the ``\texttt{instance\_of}" relation as the ontology-view concept set $\mathcal{C}_O$ on Wikidata, and we consider this ``\texttt{instance\_of}" fact set as the cross-view link set $\mathcal{H}_S$. Then, we find the tail entity set $Q$ corresponding to $\mathcal{C}_O$ through the ``\texttt{subclass\_of}" relation and update $\mathcal{C}_O$ to $\mathcal{C}_O\cup Q$ to obtain deeper concepts of the existing concepts of the ontology view. This operation is repeated until there are no deeper concepts to finish $\mathcal{C}_O$.

\begin{table}[t]
\small
\centering
\setlength{\tabcolsep}{2mm}{
\begin{tabular}{llrrr}
\toprule \multicolumn{2}{c}{\textbf{Splited Dataset}} & \textbf{$\mathcal{H}_{I}$} & \textbf{$\mathcal{H}_{O}$} &\textbf{$\mathcal{H}_{S}$} \\
\midrule 
\multirow{4.5}{*}{\textbf{JW44K-6K}}
& All & 514,822& 17,667&61,874 \\
\cmidrule{2-5}
& Train & 412,477& 14,130&49,517 \\
& Valid &51,129 &1,761 &6,178 \\
& Test &51,152 &1,774 &6,178 \\
\midrule 
\multirow{4.5}{*}{\textbf{HTDM}}
& All &208 &470 &469 \\
\cmidrule{2-5}
& Train &166 &376 &375 \\
& Valid &21 &47 &47 \\
& Test &21 &47 &47 \\
\bottomrule
\end{tabular}}
\caption{\label{T2}
Data Statistics for Dataset Partition of two DH-KG datasets, JW44K-6K and HTDM.
}
\end{table}

\begin{table}[t]
\small
\centering
\setlength{\tabcolsep}{1.5mm}{
\begin{tabular}{lccc}
\toprule \textbf{Task} & \textbf{Train} & \textbf{Valid} &\textbf{Test} \\
\midrule 
\textbf{$\mathcal{G}_I$ LP}&$\mathcal{H}_{I}^{Train}\cup \mathcal{H}_{O}^{Train} \cup \mathcal{H}_{S}^{Train}$&$\mathcal{H}_{I}^{Valid}$ &$\mathcal{H}_{I}^{Test}$ \\
\textbf{$\mathcal{G}_O$ LP}&$\mathcal{H}_{I}^{Train}\cup \mathcal{H}_{O}^{Train} \cup \mathcal{H}_{S}^{Train}$ &$\mathcal{H}_{O}^{Valid}$ &$\mathcal{H}_{O}^{Test}$ \\
\textbf{ET}&$\mathcal{H}_{I}^{Train}\cup \mathcal{H}_{O}^{Train} \cup \mathcal{H}_{S}^{Train}$ &$\mathcal{H}_{S}^{Valid}$ &$\mathcal{H}_{S}^{Test}$ \\
\bottomrule
\end{tabular}}
\caption{\label{T3}
The usage of dataset partition in instance-view LP, ontology-view LP, and ET tasks of DH-KG.
}
\end{table}

\textbf{Step3. Extract the hyper-relational facts from the two views, respectively.} 
First, we extract the hyper-relational fact set $\mathcal{H}_{I}$, where all entities and auxiliary values in extracted facts are from $\mathcal{E}_I$. Then we extract the hyper-relation $\mathcal{H}_{O}$ in the ontology layer concept set, where all concepts and auxiliary values in extracted facts are from $\mathcal{C}_O$.

\textbf{Step4. Dataset synthesis.} 
We extract relation set from $\mathcal{H}_I$, $\mathcal{H}_O$ as $\mathcal{R}_I$, $\mathcal{R}_O$, and we construct the instance view  $\mathcal{G}_I$=$\{\mathcal{E}_I, \mathcal{R}_I, \mathcal{H}_I\}$ and the ontology view $\mathcal{G}_O$=$\{\mathcal{E}_O, \mathcal{R}_O, \mathcal{H}_O\}$. Finally, we get a DH-KG dataset JW44K-6K, $\{\mathcal{G}_I,\mathcal{G}_O,\mathcal{H}_S\}$.

\subsection{HTDM Dataset}
While researching hypertension medical data, we discover that it contains many hierarchical structures suitable for DH-KG datasets. Existing medical knowledge graphs contain rich hierarchical knowledge, and patient data contains a wide range of entities. Therefore, we construct our hypertensive disease medication dataset HTDM based on DH-KG for patient medication decisions with medical knowledge by link prediction.

HTDM consists of an ontology view composed of medical knowledge and an instance view derived from actual cases, and we construct it in four steps: ontology view establishment based on medical knowledge, instance view construction based on patient data, cross-view links collection, and medicine set definition.

\subsection{Dataset Analysis}

Table~\ref{T1} presents the data statistics of the JW44K-6K, and HTDM datasets, Table~\ref{T2} depicts our division of the datasets into a training set, a validation set, and a test set in the ratio of 8:1:1,  and Table~\ref{T3} shows what datasets we use for LP and ET on DH-KG.

We can observe that JW44K and HTDM, as DH-KG datasets, contain hyper-relational facts and dual-view to represent the hierarchical structure and are more comprehensive than other datasets. The JW44K-6K dataset has a larger volume than HTDM, but HTDM is a medical dataset and contains hyper-relational data with greater arity.

\section{DHGE Model}
\label{s5}

In this section, we introduce our proposed DH-KG embedding model DHGE\footnote{\url{https://github.com/LHRLAB/DHGE}}, which consists of the intra-view GRAN encoder, cross-view association, and joint learning on DH-KG.

\subsection{Intra-view GRAN Encoder }

For intra-view hyper-relation learning, we use the GRAN~\citep{GRAN} learning method to update embedding, predict the entity or relation embedding in masked position, and calculate the intra-view loss in each view.

\begin{figure*}
    \centering
    \includegraphics[width=17.5cm]{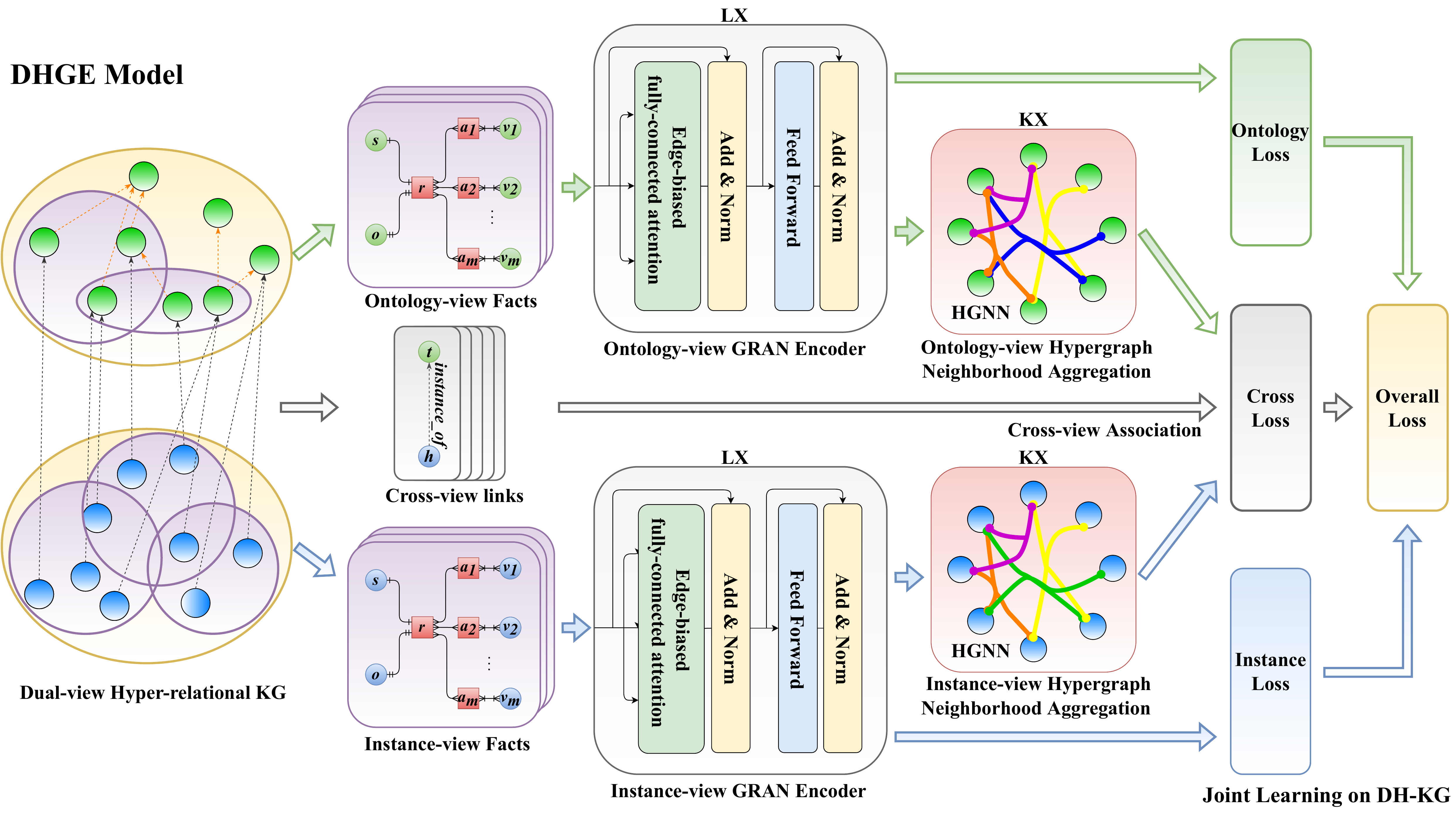}
    \caption{The DHGE model consists of three modules: GRAN encoders, HGNNs, and joint learning.}
    \label{fig:my_label}
\end{figure*}

\textbf{GRAN Learning Method.} GRAN learns a hyper-relational fact $\mathcal{H}=\left((s, r, o),\left\{\left(a_{i}: v_{i}\right)\right\}_{i= 1}^{m}\right)$ as a heterogeneous graph, then uses the mask learning strategy to construct the input of the model. The mask strategy generates (2m+3) times as many hyper-relational facts by replacing entities at each location with [MASK]. We divide these hyper-relational facts into multiple batches for training, denoting batch\_size as B. Next, GRAN uses the edge-biased fully-connected attention to learn the heterogeneous graph. We randomly embed the intra-view entity set $\mathcal{V}$ into $\mathbf{X}=(\mathbf{x_1},\ldots,\mathbf{x_{|\mathcal{V}|}})\in R^{|\mathcal{V}|\times d}$, where $d$ is the entity embedding dimension, and replace entities in all hyper-relational facts with corresponding embeddings. Taking a batch $\mathbf{S} = ({\mathbf{S_1},\ldots,\mathbf{S_B}})\in R^{B\times L\times d}$ training as an example, where L is the maximum number of entities in a 
hyper-relational fact. We take $S$ as the input to the GRAN encoder, and the updated $S$ after $l$-th layer GRAN encoder ($GRAN\_E$) is: 
\begin{equation}
\mathbf{S}^{(l)}=GRAN\_E(\mathbf{S}^{(l-1)}),
\end{equation}
where $\mathbf{S}^{(l)}\in R^{B\times L\times d}$ is the output result of the $l$-th layer $l=1,...,L$.

\textbf{Intra-view Loss. } We utilize an L-layer GRAN encoder for node embedding updates at the instance and ontology layers. After getting the updated nodes, we take out the node embedding matrix $\mathbf{h}\in R^{B\times d}$ corresponding to [MASK] in $S$, and perform an $MLP$ and a linear layer operation,
\begin{equation}
\mathbf{p}_i=MLP\left( \mathbf{h}_i \right)\mathbf{W}^T,
\end{equation}
where $\mathbf{W}\in R^{|\mathcal{V}|\times d}$ share the initial entity embedding matrix, $MLP$ is $R^{d}\rightarrow R^{d}$, $\mathbf{h}_i$ and $\mathbf{p}_i\in R^{B\times |\mathcal{V}|}$ are the $i$-th item of $\mathbf{h}$ and $\mathbf{p}$, respectively, where $\mathbf{p}$ is the similarity matrix of all [MASK] embeddings in $S$ with all entity embeddings. Finally, the cross-entropy with label smoothing between the prediction and the label is calculated as the intra-view training loss:
\begin{equation}
\mathcal{L}_{intra}=\sum_{t=1}^B{\mathbf{y}_t\log \frac{exp\left( \mathbf{p}_t \right)}{\sum_{k=1}^{|V|}{exp\left( \mathbf{p}_k \right)}}},
\end{equation}
where $\mathbf{y}_t$ is the $t$-th entry of the label $\mathbf{y}$.

Therefore, the loss obtained at the instance layer is recorded as $\mathcal{L}_{intra}^{\mathcal{G}_I}$, and the loss obtained at the ontology layer is recorded as $\mathcal{L}_{intra}^{\mathcal{G}_O}$.

\begin{table*}[t]
\small
\centering
\setlength{\tabcolsep}{0.55mm}{
\begin{tabular}{llrrrr|rrrr|rrrr|rrrr}
\toprule
\multirow{3}{*}{\textbf{Catagory}} & \multirow{3}{*}{\textbf{Model}} & \multicolumn{8}{c}{\textbf{JW44K-6K: $\mathcal{G}_I$ Link Prediction}} & \multicolumn{8}{c}{\textbf{JW44K-6K: $\mathcal{G}_O$ Link Prediction}} \\
 \cmidrule(lr){3-10} \cmidrule(lr){11-18}
 &  & \multicolumn{4}{c}{\textbf{Entity Prediction}} & \multicolumn{4}{c}{\textbf{Relation Prediction}} & \multicolumn{4}{c}{\textbf{Entity Prediction}} & \multicolumn{4}{c}{\textbf{Relation Prediction}} \\
  \cmidrule(lr){3-6} \cmidrule(lr){7-10} \cmidrule(lr){11-14} \cmidrule(lr){15-18}
 &  & \textbf{MRR} & \textbf{H@1} & \textbf{H@3} & \textbf{H@10} & \textbf{MRR} & \textbf{H@1} & \textbf{H@3} & \textbf{H@10} & \textbf{MRR} & \textbf{H@1} & \textbf{H@3} & \textbf{H@10} & \textbf{MRR} & \textbf{H@1} & \textbf{H@3} & \textbf{H@10} \\
 \midrule
\multirow{4}{*}{\textbf{ST-KGE}} & TransE & 0.197 & 0.091 & 0.248 & 0.404 & - & - & - & - & 0.094 & 0.027 & 0.135 & 0.255 & - & - & - & - \\
 & RotatE & 0.243 & 0.152 & 0.280 & 0.425 & - & - & - & - & 0.162 & 0.051 & 0.209 & 0.298 & - & - & - & - \\
 & TuckER & 0.296 & 0.221 & 0.330 & 0.435 & - & - & - & - & 0.182 & 0.077 & 0.161 & 0.278 & - & - & - & - \\
 & AttH & 0.245 & 0.170 & 0.271 & 0.384 & - & - & - & - & 0.137 & 0.056 & 0.159 & 0.309 & - & - & - & - \\
 \midrule
\multirow{3}{*}{\textbf{SH-KGE}} & HINGE & 0.227 & 0.161 & 0.255 & 0.354 & 0.849 & 0.763 & 0.930 & 0.978 & 0.113 & 0.064 & 0.137 & 0.206 & 0.591 & 0.421 & 0.722 & 0.885 \\
 & StarE & 0.314 & 0.233 & 0.350 & 0.469 & - & - & - & - & 0.150 & 0.084 & 0.169 & 0.307 & - & - & - & - \\
 & GRAN & 0.444 & 0.380 & 0.477 & 0.562 & 0.948 & 0.920 & 0.974 & 0.990 & 0.209 & 0.153 & 0.230 & 0.314 & 0.718 & 0.609 & 0.803 & 0.893 \\
 \midrule
\multirow{2}{*}{\textbf{DT-KGE}} & JOIE & 0.181 & 0.108 & 0.218 & 0.315 & - & - & - & - & 0.124 & 0.077 & 0.140 & 0.215 & - & - & - & - \\
 & HyperKA & 0.164 & 0.083 & 0.192 & 0.302 & - & - & - & - & 0.103 & 0.062 & 0.118 & 0.182 & - & - & - & - \\
 \midrule \midrule
\multirow{4}{*}{\textbf{DH-KGE}} & DH w/o GE & 0.044 & 0.022 & 0.051 & 0.081 & 0.193 & 0.091 & 0.193 & 0.414 & 0.015 & 0.005 & 0.013 & 0.030 & 0.709 & 0.595 & 0.793 & 0.856 \\
 & DH w/o HGA & 0.450 & 0.382 & 0.485 & 0.573 & 0.955 & 0.930 & 0.978 & 0.990 & 0.219 & 0.175 & 0.237 & 0.301 & 0.696 & 0.624 & 0.751 & 0.807 \\
 & DH w/o JL & 0.446 & 0.377 & 0.482 & 0.574 & 0.953 & 0.928 & 0.977 & 0.989 & 0.216 & 0.168 & 0.235 & 0.307 & 0.714 & 0.639 & 0.773 & 0.829 \\
 & DHGE & \textbf{0.453} & \textbf{0.388} & \textbf{0.486} & \textbf{0.575} & \textbf{0.958} & \textbf{0.932} & \textbf{0.982} & \textbf{0.994} & \textbf{0.227} & \textbf{0.180} & \textbf{0.242} & \textbf{0.320} & \textbf{0.745} & \textbf{0.670} & \textbf{0.807} & \textbf{0.916}\\
\bottomrule                        
\end{tabular}}
\caption{\label{t4}
Results of Link Prediction on JW44K-6K. 
}
\end{table*}

\subsection{Cross-view Association}

To complete the cross-view association, we propose using hypergraph neighborhood aggregation and cross-view loss function after learning the intra-view GRAN encoder.

\textbf{Hypergraph Neighborhood Aggregation.} Because of two or more arity hyper-relational facts, each view of DH-KG can be regarded as a hypergraph $\mathcal{G}_H=(\mathcal{V}, \mathcal{E})$ composed of entity nodes and hyper-edges. To combine the information of the two views, we use HGNN~\citep{HGNN} to aggregate the node information with hyper-edges. Through the GRAN encoder used before, we have obtained all node embedding  $\mathbf{U}^{(0)}=(\mathbf{u}_1,...,\mathbf{u}_n)=\mathbf{X}^{(L)} \in R^{|\mathcal{V}|\times d}$ as the input of HGNN. The message passing process of the HGNN from layer $(k-1)$ to layer $k$ is defined as follows:
\begin{equation}
\mathbf{W}_H=\mathbf{D}_{v}^{-1 / 2} \mathbf{H} \mathbf{D}_{e}^{-1} \mathbf{H}^{\top} \mathbf{D}_{v}^{-1/2},
\end{equation}
\begin{equation}
\mathbf{U}^{(k)}=\mathbf{U}^{(k-1)}+\sigma\left(\mathbf{W}_H\mathbf{U}^{(k-1)}\boldsymbol{\Theta}^{(k)}+\mathbf{b}^{(k)}\right),
\end{equation}
where $\mathbf{\Theta}^{(k)}\in R^{|\mathcal{V}|\times|\mathcal{V}|}$ is the transformation matrix, $\mathbf{b}^{(k)}\in R^{|\mathcal{V}|}$ is the bias vector of the $k$th layer, $\sigma$ is an activation function, $\mathbf{H}\in R^{|\mathcal{V}|\times|\mathcal{E}|}$ is the association matrix of the knowledge hypergraph,  $\mathbf{D}_v\in R^{|\mathcal{V}|}$ is the degree of the node, $\mathbf{D}_e\in R^{|\mathcal{E}|}$ is the degree of the hyperedge, and $\mathbf{U}^{(k)}\in R^{d}$ is the output result of the $k$-th layer $k=1,...,K$. We combine the input and output representations $\mathbf{U}=\mathbf{U}^{(0)}+\mathbf{U}^{(K)}$ as the final embedding to further benefit from relational transformation

\textbf{Cross-view Loss. } We obtain $\mathbf{U}^{\mathcal{G}_{I}}$ and $\mathbf{U}^{\mathcal{G}_{O}}$ after Hypergraph Neighborhood Aggregation on the instance layer and the ontology layer, respectively. According to the cross-view link set $H_S$, the embedding of the head entity $h_{S}$ is in the embedding space of the instance view denoted as $\mathbf{U}^{\mathcal{G}_{I}}({h_{S}})$, and the embedding of the tail entity $t_S$  is in the ontology view embedding space denoted as $\mathbf{U}^{\mathcal{G}_{O}}({t_{S}})$. We map the head entity set to the same ontology view embedding space as the tail entity through the mapping operation:
\begin{equation}
f(\mathbf{U}^{\mathcal{G}_{I}}({h_{S}}))=\mathbf{W}_{map}(\mathbf{U}^{\mathcal{G}_{I}}({h_{S}}))+\mathbf{b}_{map}, 
\end{equation}
and design the cross-link loss function as follows:
\begin{equation}
\begin{split}
\mathcal{L}_{\text{Cross}}=&\frac{1}{\left|\mathcal{H}_{\mathcal{S}}\right|}\sum_{\substack{\left(h_{S}, t_{S}\right) \in \mathcal{H}_{\mathcal{S}}\wedge\left(h_{S}, t_{S}^{\prime}\right) \notin \mathcal{H}_{\mathcal{S}}}}\\
&[\gamma+\left\|f\left(\mathbf{U}^{\mathcal{G}_{I}}({h_{S})}\right)-\mathbf{U}^{\mathcal{G}_{O}}({t_{S})}\right\|_{2}\\
&-\left\|f\left(\mathbf{U}^{\mathcal{G}_{I}}({h_{S})}\right)-\mathbf{U}^{\mathcal{G}_{O}}({t_{S}^{\prime})}\right\|_{2} ],
\end{split}
\end{equation}
where we use the 2-norm function to calculate the distance deviation of entities and concepts under the same embedding space, $t_{S}^{\prime}$ is the negative example of $t_{S}$, and $\gamma$ is a parameter of margin.

\subsection{Joint Learning on DH-KG}

We combine the training losses of the instance view, ontology view, and cross-view link set to get a joint learning loss:
\begin{equation}
\mathcal{L}_{\text {overall}}=\mathcal{L}_{\text {Intra }}^{\mathcal{G}_{I}}+\mathcal{L}_{\text {Intra }}^{\mathcal{G}_{O}}+\omega \cdot \mathcal{L}_{\text {Cross }},
\end{equation}
and we adopt the Adam optimizer to optimize the three losses alternately, where $\omega$ distinguishes the learning rate of intra-view loss and cross-view loss to achieve joint learning on DH-KG.

\section{Experiments}
\label{s6}
In this section, we evaluate our proposed model DHGE with the tasks of link prediction (LP), entity typing (ET), and hypertension disease medicine prediction (MP, MCP), respectively. We introduce our experimental setup, followed by experimental results.
\subsection{Experimental Setup}
\label{6.1}
\subsubsection{Datasets}
Our experiments are done on two DH-KG datasets, JW44K-6K and HTDM mentioned above. On JW44K-6K, we evaluate the LP and ET tasks on DH-KGs. On HTDM, we only perform entity prediction of the object position in the instance view to simulate the prediction of the doctor's medication for real hypertensive patients. 

\subsubsection{Baselines}
We compare DHGE with the following representative KG embedding (KGE) methods. The first category is Single-view Triple-based KGE Methods (ST-KGE): \textbf{TransE}~\citep{TransE}, \textbf{RotatE}~\citep{RotatE}, \textbf{TuckER}~\citep{TuckER}, and \textbf{AttH}~\citep{AttH}, which ignore the auxiliary attribute-value pairs of the hyper-relational facts and treat the dual view as a single view for the experiment. The second category is Single-view Hyper-relational KGE Methods (SH-KGE): \textbf{HINGE}~\citep{HINGE}, \textbf{StarE}~\citep{StarE}, and \textbf{GRAN}~\citep{GRAN}, which treats facts as a hyper-relational paradigm in a single view. The third category is Dual-view Triple-based KGE Methods (DT-KGE): \textbf{JOIE}~\citep{JOIE}, and \textbf{HyperKA}~\citep{HyperKA}, which can learn dual-view information, but facts are also triple-based.

\subsubsection{Ablations}
We ignore the three modules of DHGE: intra-layer GRAN encoder, HGNN neighbor information aggregation, and joint learning, respectively, to form three ablation models: \textbf{DHGE w/o GE}, \textbf{DHGE w/o HGA}, and  \textbf{DHGE w/o JL} for ablation study.

\subsubsection{Evaluation Metrics}
For the task of LP and ET, we choose mean reciprocal rank (MRR) and hits at 1, 3, and 10 (H@1, H@3, H@10) evaluation indicators introduced by \citet{TransE}. For Medicine Prediction (MP, MCP) on HTDM, we cancel the H@10 indicator for the requirement of high accuracy.

\subsubsection{Hyperparameters and Enviroment}
Since hyperparameters are sensitive to experimental results, we use the following hyperparameters for all models: embedding dimension = 256 for both entity and relation, epoch = 100, learning rate = 5e$-$4, and batch size = 256. All experiments were conducted with a single 11GB GeForce GTX 1080Ti GPU, and all results were obtained by averaging 3 experiments.

\subsection{Main Results and Analysis}
\label{6.2}
In this experiment, we compare the LP and ET results of DHGE with all baselines and ablations. 

\subsubsection{Link Prediction}
We have defined the LP task on DH-KG, which includes entity and relation prediction for instance and ontology views. The conclusions are drawn from comparing DHGE to baselines and ablations in Table~\ref{t4}.

\textbf{Comparison with ST-KGE:} DHGE outperforms the  ST-KGE model in JW44K-6K LP by a large margin because triple-based embedding methods are inherently unable to model neither auxiliary attribute-value pairs of hyper-relational facts nor hierarchical information.
\textbf{Comparison with SH-KGE:} The intra-view LP task results of DHGE are generally higher than that of SH-KGE because cross-view links influence the embedding in one view and guide the embedding in another to be more accurate.
\textbf{Comparison with DT-KGE:} The intra-view LP task performance of JOIE and HyperKA are scanty, even worse than the ST-KGE models because they are triple-based KGE methods and focus more on the ET task across views.
\textbf{Comparison with DH-KGE ablation model:} The performance of all ablations degrades, and DHGE w/o GE is the worst performer, indicating that GRAN Encoder, Hypergraph Neighbourhood Association, and Joint Learning are useful in LP, and GRAN Encoder plays the most critical role.
\textbf{Comparison between instance view and ontology view results:} The LP task performance of the ontology view is much more deficient than that of the instance view, indicating that the LP of the ontology view is more complicated because it is difficult for humans to predict and understand the facts between concept entities.

\begin{table}[t]
\small
\centering
\setlength{\tabcolsep}{1.5mm}{
\begin{tabular}{llrrrr}
\toprule
\multirow{2.5}{*}{\textbf{Catagory}} & \multirow{2.5}{*}{\textbf{Model}} & \multicolumn{4}{c}{\multirow{1}{*}{\textbf{JW44K-6K: Entity Typing}}} \\
 \cmidrule(lr){3-6}
 &  & \multicolumn{1}{r}{\textbf{MRR}} & \multicolumn{1}{r}{\textbf{H@1}} & \multicolumn{1}{r}{\textbf{H@3}} & \multicolumn{1}{r}{\textbf{H@10}} \\
\midrule
\multirow{4}{*}{\textbf{ST-KGE}} & TransE & 0.085 & 0.031 & 0.102 & 0.204 \\
 & RotatE & 0.362 & 0.293 & 0.390 & 0.495 \\
 & TuckER & 0.405 & 0.350 & 0.429 & 0.501 \\
 & AttH & 0.416 & 0.348 & 0.448 & 0.539 \\
\midrule
\multirow{3}{*}{\textbf{SH-KGE}} & HINGE & 0.109 & 0.062 & 0.132 & 0.198 \\
 & StarE & 0.380 & 0.321 & 0.404 & 0.479 \\
 & GRAN & 0.422 & 0.343 & 0.460 & 0.571 \\
\midrule
\multirow{2}{*}{\textbf{DT-KGE}} & JOIE & 0.603 & 0.535 & 0.645 & 0.773 \\
 & HyperKA & \textbf{0.692} & 0.626 & \textbf{0.736} & 0.798 \\
\midrule\midrule
\multirow{4}{*}{\textbf{DH-KGE}} & DHGE w/o GE & 0.492 & 0.380 & 0.560 & 0.695 \\
 & DHGE w/o HGA & 0.662 & 0.627 & 0.675 & 0.702 \\
 & DHGE w/o JL & 0.428 & 0.332 & 0.468 & 0.613 \\
 & DHGE & 0.690 & \textbf{0.637} & 0.726 & \textbf{0.805}\\
\bottomrule
\end{tabular}}
\makeatletter\def\@captype{table}\makeatother\caption{\label{t5}
Results of Entity Typing on JW44K-6K. 
}
\end{table}


\subsubsection{Entity Typing}
In this paper, we define the ET task on DH-KG, that is, to correspond the concepts in the ontology view with the entities in the instance view. We compare DHGE with baselines and ablations in Table~\ref{t5} and obtain the following conclusions.

\textbf{Comparison with ST-KGE and SH-KGE:} These two KGE methods use the single view for entity prediction tasks. In the ET task, DHGE with cross-view learning performs better than a single-view model.
\textbf{Comparison with DT-KGE:} The performance of models of DT-KGE is close to that of DHGE because JW44K-6K has rich hierarchical information. DT-KGE models still have good performance though they are triple-based.
\textbf{Comparison with DH-KGE ablation model:} The ablation model without joint training has the worst performance in the entity typing task, indicating that joint training is the most critical part of the ET task because of rich cross-view links on DH-KG.

\subsubsection{Hypertension Medicine Prediction}
Here, we present an example of a medical application of DH-KGs: hypertension medication prediction. We define the 133 entities of medicine as a fine-grained medicine set and the 8 entities of medicine class as a coarse-grained medicine class set. Then, we define the prediction within these two sets as a medicine prediction task (MP) and a medicine class prediction task (MCP), respectively. These two experiments do the LP task on the hyper-relational tail entity position of the instance view through the joint learning of the DH-KG. 

Table~\ref{t6} shows that the prediction results of HTDM are consistent with the instance-view link prediction when comparing the DHGE model with the baseline and ablation model. The MP and MCP results of DHGE are significantly higher than baselines and ablations. DHGE’s MCP Hits@1 can reach 0.667, and Hits@3 can reach 0.905, indicating the excellent application prospect of DHGE.  

\begin{table}[t]
\small
\centering
\setlength{\tabcolsep}{0.5mm}{
\begin{tabular}{llrrr|rrr}
\toprule
\multirow{2.5}{*}{\textbf{Catagory}} & \multirow{2.5}{*}{\textbf{Model}} & \multicolumn{3}{c}{\multirow{1}{*}{\textbf{HTDM: MP}}} & \multicolumn{3}{c}{\multirow{1}{*}{\textbf{HTDM: MCP}}} \\
  \cmidrule(lr){3-5} \cmidrule(lr){6-8}
 &  & \multicolumn{1}{r}{\textbf{MRR}} & \multicolumn{1}{r}{\textbf{H@1}} & \multicolumn{1}{r}{\textbf{H@3}} & \multicolumn{1}{r}{\textbf{MRR}} & \multicolumn{1}{r}{\textbf{H@1}} & \multicolumn{1}{r}{\textbf{H@3}} \\
\midrule
\multirow{4}{*}{\textbf{ST-KGE}} & TransE & 0.041 & 0.006 & 0.047 & 0.224 & 0.142 & 0.428 \\
 & RotatE & 0.074 & 0.023 & 0.095 & 0.469 & 0.180 & 0.519 \\
 & TuckER & 0.213 & 0.075 & 0.214 & 0.579 & 0.419 & 0.642 \\
 & AttH & 0.170 & 0.071 & 0.190 & 0.526 & 0.357 & 0.576 \\
\midrule
\multirow{3}{*}{\textbf{SH-KGE}} & HINGE & 0.189 & 0.084 & 0.189 & 0.624 & 0.571 & 0.642 \\
 & StarE & 0.224 & 0.095 & 0.285 & 0.637 & 0.571 & 0.667 \\
 & GRAN & 0.374 & 0.238 & 0.428 & 0.679 & 0.633 & 0.857 \\
\midrule
\multirow{2}{*}{\textbf{DT-KGE}} & JOIE & 0.073 & 0.014 & 0.090 & 0.361 & 0.181 & 0.514 \\
 & HyperKA & 0.062 & 0.013 & 0.072 & 0.323 & 0.119 & 0.461 \\
\midrule\midrule
\multirow{4}{*}{\textbf{DH-KGE}} & DH w/o GE & 0.274 & 0.095 & 0.381 & 0.559 & 0.285 & 0.809 \\
 & DH w/o HGA & 0.306 & 0.238 & 0.333 & 0.668 & 0.476 & 0.857 \\
 & DH w/o JL & 0.301 & 0.190 & 0.333 & 0.628 & 0.428 & 0.761 \\
 & DHGE & \textbf{0.445} & \textbf{0.381} & \textbf{0.572} & \textbf{0.772} & \textbf{0.667} & \textbf{0.905}\\
\bottomrule
\end{tabular}}
\makeatletter\def\@captype{table}\makeatother\caption{\label{t6}
Results of Medicine Prediction on HTDM. 
}
\end{table}


\section{Conclusion}
\label{s7}
This paper focuses on dual-view hierarchical hyper-relational embedding. We introduce the tasks of link prediction and entity typing for DH-KGs and create two new datasets, JW44K-6K and HTDM. Our DHGE model captures hyper-relational and hierarchical information and outperforms related models. In addition, we also demonstrate a use case in hypertension medication.

\section*{Acknowledgements}
This work is supported by the National Science Foundation of China (Grant No. 62176026, Grant No. 61902034) and Beijing Natural Science Foundation (M22009, L191012). This work is also supported by the BUPT Postgraduate Innovation and Entrepreneurship Project led by Haoran Luo.

\bibliography{aaai23}

\end{document}